%% file: main.tex
\documentclass{article}
\usepackage{spconf,amsmath,graphicx}
\usepackage{enumitem}
\usepackage{makecell}

\title{Impact of multiple modalities on emotion recognition: investigation into 3d facial landmarks, action units, and physiological data}

\name{Diego Fabiano, Manikandan Jaishanker, Shaun Canavan}
\address{University of South Florida}
\begin{document}
\maketitle

\input{abstract.tex}


\input{introduction.tex}
\input{method.tex}
\input{experimentsResults.tex}
\input{discussion.tex}
\input{ack.tex}
\input{references.tex}

\end{document}

%% file: abstract.tex
\begin{abstract}
 To fully understand the complexities of human emotion, the integration of multiple physical features from different modalities can be advantageous. Considering this, we present an analysis of 3D facial data, action units, and physiological data as it relates to their impact on emotion recognition. We analyze each modality independently, as well as the fusion of each for recognizing human emotion. This analysis includes which features are most important for specific emotions (e.g. happy). Our analysis indicates that both 3D facial landmarks and physiological data are encouraging for expression/emotion recognition. On the other hand, while action units can positively impact emotion recognition when fused with other modalities, the results suggest it is difficult to detect emotion using them in a unimodal fashion.
 \end{abstract}

%% file: introduction.tex
\section{Introduction}
Recognizing emotion is considered one of the most important parts of human intelligence \cite{picard2001toward} and it has applications in fields as varied as entertainment, transportation, medicine and health, and psychology. Due to this, there has been a great deal of research into human emotion recognition in the past decades, where many important advances have been made. This is due in part to the new availability of large, varied, and challenging datasets \cite{cosker2011facs, fanelli20103, koelstra2011deap, mckeown2011semaine, soleymani2011multimodal, stratou2012exploring, wang2010natural, yin126high}.

There is a large and varied body of work into facial expression recognition. Using a Spatio-Temporal Hidden Markov Model (HMM), the intra- and inter-frame information can be used for this task \cite{SunTrackingVertFlow}. It has been shown that using a random forest \cite{breiman2001random} along with a Deformation Vector Field \cite{drira20123d}, to obtain the local deformations of the face over time, can be used to accurately classify expressions. Facial expressions have also been successfully classified using a Support Vector Machine (SVM) with a radial basis function (RBF) kernel with geometrical coordinates, as well as the normal of the coordinates \cite{fang20123d}. Lucey et al \cite{lucey2010extended}, analyzed videos of patients with shoulder injuries to automatically recognize pain. In this work, an Active Appearance Model \cite{cootes2001active} was used to detect Facial Action Units to distinguish pain on facial expressions. They detail 84.7 for area under the ROC curve on the UNBC-McMaster Shoulder Pain Database \cite{prkachin2008structure}. This study is encouraging as it suggests Action Units can be used to recognize emotions (e.g. pain).

Deep learning has shown recent success in expression recognition. Using a Boosted Deep Belief Network, Liu et al. \cite{liu2014facial} trained feature learning, selection, and classifier construction iteratively in a unified loopy framework; which showed an increase in the classification accuracy. De-expression Residue Learning \cite{yang2018facial} was also proposed which can generate a corresponding neutral expression given an arbitrary facial expression from an image. Yang et al. \cite{yang2018identity} proposed regenerating expression from input facial images. By using a conditional GAN \cite{mirza2014conditional}, they developed an identity adaptive feature space that can handle variations in subjects. 

Facial expression recognition is a popular approach to recognizing emotion, however, there is also a varied body of work that makes use of multimodal data for emotion recognition. Soleymani et al. \cite{soleymani2011multimodal} incorporated electroencephalogram, pupillary response, and gaze distance information from 20 videos. They used this data to train an SVM to classify arousal and valence for 24 participants. Kessous also showed an increase of more than 10\% when using a multimodal approach \cite{kessous2010multimodal}. They used a Bayesian classifier, and fused facial expression with speech data that consisted of multiple languages including Greek, French, German, and Italian. 

While these works, and others, have had success detecting expressions and emotion with multimodal data, little work has been done on analyzing their impact on recognition. Motivated by this, we present an analysis of multimodal data and the impact each modality has on emotion recognition. Our contributions can be summarized as follows:

\begin{enumerate}[topsep=0pt,itemsep=-1ex,partopsep=1ex,parsep=1ex]
    \item A detailed analysis of physiological data, 3D landmarks, and facial action units (AU) \cite{FACS} both independently and combined at the feature level (unimodal vs. multimodal), for emotion recognition, is presented.
    \item Insight is provided on the impact of physiological data, 3D landmarks, and AUs for positively influencing emotion recognition studies.
    \item To the best of our knowledge, this is the first work to conduct this type of analysis on the BP4D+ multimodal dataset \cite{zhang2016multimodal}, resulting in a baseline for future analyses.
\end{enumerate}

%% file: method.tex
\section{Data Selection and Feature Extraction}
We propose to use 3D facial data (landmarks), action units and physiological data in our analysis. We chose these 3 modalities based on their complementary nature. First, given movement, and the shape of the face changes (3D landmarks), we can also assume that there will be a change in the occurrence of action units \cite{ekman1978Facs}. We have also chosen the complementary modality, physiological data, as facial expressions can be faked. It has been observed that people smile during negative emotional experiences \cite{ekman1989argument}. Considering this, physiological data can complement the other two modalities for recognizing emotion.

To conduct the proposed analysis, a suitably large corpus of emotion data is needed that contains 3D facial data, action units, and physiological data. For our experiments we have chosen the BP4D+ multimodal spontaneous emotion corpus \cite{zhang2016multimodal}. In total, there are over 1.5 million frames of multimodal available in the BP4D+. For this study we use 192,452 frames of multimodal data from all 140 subjects. This subset of data contains 4 target emotions that are happiness, embarrassment, fear, and pain. We are using this subset, as it is largest set of frames, in BP4D+, that are encoded with action units.

\subsection{3D facial data}
\begin{figure}
\includegraphics[width=8.5cm, height=4.5cm]{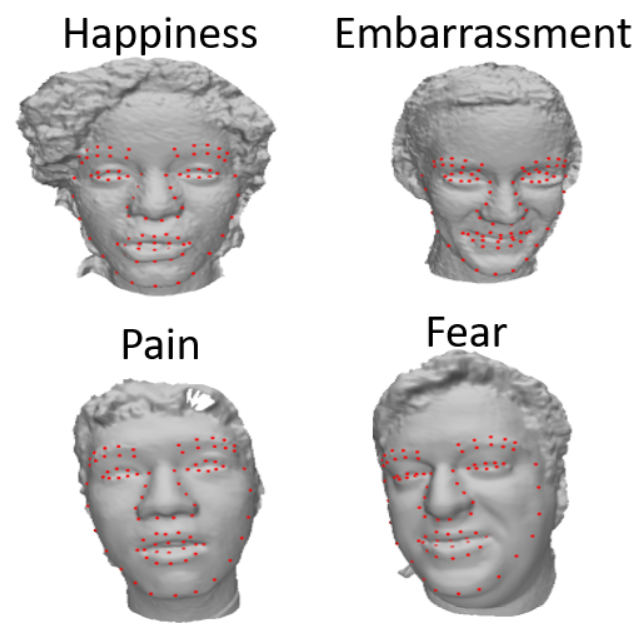}
\caption{3D facial landmarks on corresponding 3D mesh model for our targeted emotions of happiness, embarrassment, pain, and fear from the BP4D+ \cite{zhang2016multimodal}.}
\label{fig:3dLandmarks}
\end{figure}

For our study we used 83 3D facial landmarks (same as seen in BP4D+) to represent the face. Each of the landmarks were detected using a shape index-based statistical shape model (SI-SSM) \cite{canavan2015landmark}, that creates shape index-based patches from global and local features of the face. These global and local features are concatenated into one model, which is then used along with a cross-correlation matching technique to match the training data to an input mesh model. Examples of detected 3D facial landmarks can be seen in Fig. \ref{fig:3dLandmarks}. For our 3D facial data feature vector, we directly use the coordinates  (x, y, z) of the 3D tracked facial landmarks as they can accurately represent the induced expression that can be seen in the entire 3D model \cite{fabianoICIP2018}, which contains approximately 30k-50k vertices; where our reduced feature vector contains 249 features (83 – 3D coordinates). Using this reduced feature space (relative to the entire 3D mesh) allows for lower dimensional data, without sacrificing any recognition accuracy.

\subsection{Action units}
For each of the 4 emotions that have action units coded, a total of 35 action units (AUs) were coded by five different expert FACS coders. For each task of all 140 subjects approximately 20 seconds of the most expressive part of the sequence was annotated, giving us our 192,452 frames of multimodal data that we use for our study. For our AU feature vector, we include the occurrence of all annotated AUs for each frame where 1 corresponds to the AU being present and 0 corresponds to the AU not being present in the current frame. There are some instances in the BP4D+ where the AU occurrence is listed as 9, which is referred to as unknown. For our experiments, 9 is treated as a 0 (i.e. not present). 

\subsection{Physiological data}
For each subject and task, the BP4D+ contains 8 separate measurements of physiological data derived from blood pressure (BP), heart rate (HR), respiration (RESP), and skin conductivity (EDA). All physiological data was sampled at 1000 Hz which required us to synchronize with the available 3D facial data and corresponding action units to have accurate readings for each frame of data.  
To synchronize this, we first divide the total number of frames of physiological data by the total number of frames of 3D facial data for that task (average sync value). We then use the average value over the average sync value as our new frame. For example, given a task with 1000 frames of 3D facial data, along with 40,000 frames of diastolic BP we would have $40,000/1000=40$, resulting in us taking the average diastolic BP for every 40 frames. Calculating the mentioned average over all 40,000 frames, results in 1000 frames of diastolic BP matching to the 1000 frames of corresponding 3D facial data. In this same task, there are 400 frames that include both 3D facial landmarks and AUs (frames labeled with task, subject, and frame number). We then use the corresponding frame number to extract that exact index from the calculated diastolic BP averages. This gives us our resulting 400 frames of synchronized 3D facial data, physiological data, and action units. For our physiological feature vector, we take the average value of each frame over all eight of the data types (i.e. fuse the signals).

%% file: experimentsResults.tex
\section{Experimental Design and Results}
\subsection{Feature analysis}
A main contribution of this work is analyzing which modality and features are most important for our 4 target emotions. To do this we used principal component analysis (PCA) for feature selection keeping 95\% of the original variance. We did this for each of our unimodal feature vectors for all the training data, as well as each individual emotion. This was done to analyze which features are important for emotion recognition in a general sense, and for each targeted emotion resulting in a total of 15 total rankings (3 feature vectors for each: happy, embarrassment, pain, fear, and all emotions). The features were then ranked based on highest eigenvalue.

\begin{figure}
\includegraphics[width=8.5cm, height=4.5cm]{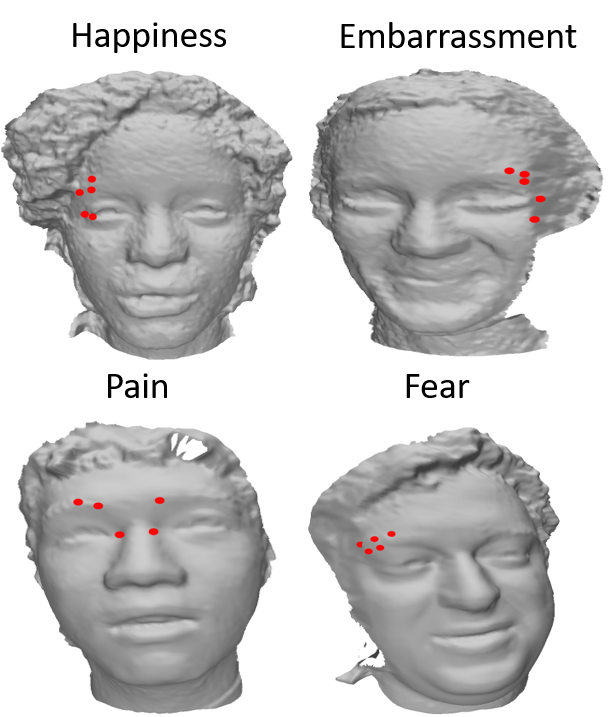}
\caption{Top 5 ranked 3D facial features across the 4 emotions.}
\label{fig:top5}
\end{figure}

\textbf{Action Units.} The top ranked action units included the lips, cheeks, nose, and eye/eyebrow regions. Across each of the target emotions, along with all combined emotions the ranked AUs were similar. The difference being their rankings change across different emotion (e.g. AU12 was ranked first for happy, while AU12 was ranked second for embarrassed). Table \ref{table:PCA}, second column, shows the top 5 ranked AUs. As can be seen here the top AUs for ‘Happy’ are 12, 6, 11, and 7. When considering the Emotion Facial Action Coding System \cite{friesenEmFacs}, which only looks at emotion-related facial action, ‘Happy’, is 6+12. This shows a correlation between the PCA rankings and the action units associated with the emotion. We also calculated the normalized AU distribution across each target emotion. This showed that while each emotion had similar occurring action units, they varied in distribution, which contributes complimentary information to the other modalities. This can explain the increase in accuracy when a multimodal approach is used (Table \ref{table:multimodal}).

\textbf{Physiological Data.} Most of the top ranked features for physiological data were variations on blood pressure (e.g. diastolic and systolic). Pulse rate was also ranked as a top feature for each of the target emotions, however, when all emotions were included in the training data, pulse rate was replaced by EDA. This suggests that skin conductivity is important for recognizing multiple emotions. It is interesting to note that for each of the 4 target emotions, not only were the top ranked features the same, they were also ranked in the same order. Although each emotion had the same ranked physiological data, they all had large variations in the data between them. This variance in data allows for a high level of recognition accuracy (Table \ref{table:unimodal}). Table \ref{table:PCA}, third column, shows the top 5 ranked physiological signals.

\textbf{3D Facial Data.} When analyzing the 3D facial data, each of the target emotions show variance in the regions of the face that were ranked for the top features. For example, happy targeted the right eye and eyebrow, and pain was across the right eyebrow, nose, and left eyebrow. These regions of the face are also consistent with the AUs ranked as the top features (e.g. mouth, face, eyes/eyebrows). See Table \ref{table:PCA} for the top 5 ranked 3D facial landmarks (face region for each) and Fig. \ref{fig:top5} for examples of these landmarks on corresponding 3D mesh models. It can be seen, in Fig. \ref{fig:top5}, that emotional variance is conveyed in different 3D regions of the face for each of the target emotions.

\begin{table}
    \caption{PCA rankings for each feature for each individual emotion along with all 4 target emotions, shown in ranked order. NOTE: number in parentheses in column four corresponds to total number of landmarks in that region.}
    \label{table:PCA}
    \centering 
    \resizebox{8.5cm}{!}{
    \begin{tabular}{|c|c|c|c|}
        \hline
        \bfseries Emotion & \bfseries \bfseries Action Units & \bfseries Phys& \bfseries \makecell{ 3D Facial \\ Landmarks}\\
        \hline
         & Lip corner puller (12) & Mean BP &  \\
         & Cheek raiser (6) & Diastolic BP &  \\
        \bfseries Happy & Upper lip raiser (10) & Raw BP & Right eye (2) \\
         & Nasolabial deepener (11) &Pulse Rate & Right eyebrow (3) \\
         & Lid tightener (7) &  &  \\
        \hline
        
        & Cheek raiser (6) & Mean BP &  \\
        & Lip corner puller (12) & Diastolic BP & \\
        \bfseries Embarrassed & Upper lip raiser (10) & Systolic BP & Left face contour (2) \\
        & Lid tightener (7) & Raw BP & Left eyebrow (3) \\
        & Nasolabial deepener (11) & Pulse Rate &  \\
        \hline 
        
        & Lip Corner Puller (12) & Mean BP & \\
        & Cheek raiser (6) & Diastolic BP & Right eyebrow (2)\\
        \bfseries Pain & Upper lip raiser (10) & Systolic BP & Nose (2) \\
        & Nasolabial deepener (11) & Raw BP & Left eyebrow(1)\\
        & Lid tightener (7) & Pulse Rate & \\
        \hline 
        
        & Upper lip raiser (10) & Mean BP & \\
        & Cheek raiser (6) & Diastolic BP & \\
        \bfseries Fear & Lid tightener (7) & Systolic BP & Right eyebrow (5)\\
        & Lip corner puller (12) & Raw BP & \\
        & Nasolabial deepener (11) & Pulse Rate & \\
        \hline 
        
        & Lip corner puller (12) & Mean BP & \\
        & Upper lip raiser (10) & Diastolic BP & \\
        \bfseries All & Cheek raiser (6) & Systolic BP & Left eyebrow (5)\\
        & Lid tightener (7) & Raw BP & \\
        & Nasolabial deepener (11) & EDA & \\
        \hline
    \end{tabular}}
\end{table}

\subsection{Emotion recognition}
To conduct our emotion recognition experiments, we created a feature vector for each unimodal and multimodal configuration (Tables \ref{table:unimodal} and \ref{table:multimodal}). We then used each of these feature vectors to train a random forest  \cite{breiman2001random} for recognizing the four target emotions. Random forests have successfully been used in a wide variety of classification tasks such as classifying ecological data \cite{cutler2007random}, real-time hand gesture recognition \cite{zhao2012real}, and head pose estimation \cite{fanelli2011real}, which makes them a natural fit for our analysis.
\begin{table}
    \caption{Unimodal emotion recognition from BP4D+.}
    \label{table:unimodal}
    \centering 
    \setlength{\tabcolsep}{3pt}
    \begin{tabular}{|c|c|c|c|}
    \hline
     & \bfseries 3D Facial Data & \bfseries Action Units & \bfseries Physiological \\
     \hline
    \bfseries Accuracy & 99.29\% &  61.94\% & \bfseries 99.94\%\\
    \hline
    \bfseries Recall & 98.80\% & 60.35\% & \bfseries 99.95\% \\
    \hline
    \bfseries Precision & 99.33\% & 61.00\%& \bfseries 99.95\%\\
    \hline
    \end{tabular}
\end{table}

\begin{table}
    \caption{Multimodal emotion recognition from BP4D+.}
    \label{table:multimodal}
    \centering 
    \resizebox{8.5cm}{!}{
    \begin{tabular}{|c|c|c|c|c|}
    \hline
     & \bfseries 3D Facial Data & \bfseries Action Units & \bfseries 3D Facial Data & \bfseries 3D Facial Data \\
    & \bfseries Action Units & \bfseries Physiological & \bfseries Physiological & \bfseries Action Units \\
    & & & & \bfseries Physiological \\
     \hline
    \bfseries Accuracy & 99.53\% & \bfseries 99.95\% &  99.76\% & 99.83\%\\
    \hline
    \bfseries Recall & 99.58\% & \bfseries 99.95\% & 99.75\% & 99.83\% \\
    \hline
    \bfseries Precision & 99.52\% & \bfseries 99.95\% & 99.75\% & 99.85\% \\
    \hline
    \end{tabular}
    }
\end{table}
\begin{table}
    \caption{Confusion matrix of action units.}
    \label{table:ConfMatAU}
    \centering 
    \setlength{\tabcolsep}{3pt}
    \begin{tabular}{|c|c|c|c|c|}
    \hline
     & Happy & Embarrassment & Fear & Pain \\
     \hline
     Happy & \bfseries 32511 & 7730 & 3373 & 7917 \\
     \hline
     Embarrassment & 17561 & \bfseries 26038 & 3238 & 5282 \\
     \hline
     Fear & 8773 & 5206 & \bfseries 14652 & 8163\\
     \hline
     Pain & 1983 & 2334 & 1685 & \bfseries 46006\\ 
     \hline
    \end{tabular}
\end{table}
\textbf{Unimodal vs. Multimodal Emotion Recognition.} We used 10-fold cross validation for each of our experiments. The results for unimodal and multimodal emotion recognition can be seen in Tables \ref{table:unimodal} and \ref{table:multimodal} respectively. When physiological data was used, recognition accuracy was highest for both unimodal and multimodal approaches, achieving an accuracy of 99.94\% for the 4 target emotions, with a unimodal approach. This result is intuitive as physiological signals are closely tied to human emotion \cite{knapp2011physiological, koelstra2011deap}. When AUs were combined with physiological data we achieved our highest recognition accuracy of 99.95\%. This also agrees with the literature that the fusion of multimodal data, including action units, can provide complimentary information and increase recognition accuracy \cite{corneanu2016survey}. Although emotion recognition from AUs shows promising results, especially when fused with other modalities, they exhibit the lowest classification rate of the unimodal feature vectors with a recognition accuracy of 61.94\%. The confusion matrices for AUs, physiological data, and AUs combined with physiological data are shown in Tables \ref{table:ConfMatAU}, \ref{table:ConfMaPhys}, and \ref{table:ConfMatAUPhys} respectively.

Fusing multimodal data has been found to increase emotion recognition including pain in infants \cite{zamzmi2016machine}. Our results show similar results with pain as well, increasing from 99.92\% with physiological data to 99.98\% when AUs were fused with physiological data. It is interesting to note, that while the overall recognition accuracy was higher when AUs were combined with physiological data, the recognition rates for both happy and fear decreased to 99.94\% and 99.90\% respectively. This can be attributed to some redundant action unit patterns between happy and fear.

\begin{table}
    \caption{Confusion matrix of physiological data.}
    \label{table:ConfMaPhys}
    \centering 
    \setlength{\tabcolsep}{3pt}
    \begin{tabular}{|c|c|c|c|c|}
    \hline
     & Happy & Embarrassment & Fear & Pain \\
     \hline
     Happy & \bfseries 51512 & 10 & 5 & 4 \\
     \hline
     Embarrassment & 21 & \bfseries 52080 & 4 & 14 \\
     \hline
     Fear & 4 & 7 & \bfseries 36780 & 3 \\
     \hline
     Pain & 22 & 13 & 6 & \bfseries 51967\\ 
     \hline
    \end{tabular}
\end{table}

\begin{table}
    \caption{Confusion matrix of action units and physiological.}
    \label{table:ConfMatAUPhys}
    \centering 
    \setlength{\tabcolsep}{3pt}
    \begin{tabular}{|c|c|c|c|c|}
    \hline
     & Happy & Embarrassment & Fear & Pain \\
     \hline
     Happy & \bfseries 51504 & 10 & 5 & 4 \\
     \hline
     Embarrassment & 10 & \bfseries 52100 & 3 & 6 \\
     \hline
     Fear & 14 & 16 & \bfseries 36758 & 6 \\
     \hline
     Pain & 3 & 9 & 1 & \bfseries 51995 \\ 
     \hline
    \end{tabular}
\end{table}

%% file: discussion.tex
\section{Discussion}
We have presented an analysis on impact of 3D facial landmarks, action units, and physiological data for emotion recognition. We have conducted experiments in both a unimodal and multimodal capacity on four target emotions. Our analysis has shown that 3D facial data shows variations in facial regions allowing for accurate emotion recognition. We have also shown that physiological data can be used for emotion recognition due to the changes across emotion. The occurrence of action units shows differences in distribution over 35 AUs across the four-target emotions, which allows for complimentary information to be used when fusing the AUs with other modalities at the feature level. Although the fusion of AUs is shown to increase the accuracy across the four tested emotions, the results also show that directly using AU occurrences without fusing other modalities, for emotion recognition, is still a challenging problem. These results suggest more research is needed to determine the positive impact of using action units in a unimodal approach for emotion recognition.

While these results are encouraging, there are some limitations to the study. First, more multimodal databases need to be investigated, as our study only made use of BP4D+. Secondly, more details are needed as to why the fusion of AU occurrences showed an increase in accuracy, while using them in a unimodal capacity generated a relatively low accuracy. Lastly, our current study only focused on four emotions due to the limited number of available action units. A much larger range of emotions are needed to fully test the efficacy of the proposed approach. Considering this, for our future work, we will detect action units \cite{baltruvsaitis2016openface}, across a larger set of data, as well as use deep neural networks and other fusion methods including score level fusion, and the fusion of deep and hand-crafted features \cite{hindujafusion}. We will also test on a larger set of multimodal datasets, and we will investigate the impact of both AU occurrences \textit{and} intensities for emotion recognition. These experiments will be conducted across a larger set of emotions that include, but are not limited to, surprise, sadness, anger, and disgust. Along with these emotions, we will also investigate subject self-reporting of emotion (i.e. perceived emotion).

%% file: ack.tex
\section*{Acknowledgment}
This material is based on work that was supported in part by an Amazon Machine Learning Research Award.

%% file: references.tex
\bibliographystyle{ieee}
\bibliography{references}

%% file: main.bbl
\begin{thebibliography}{10}\itemsep=-1pt

\bibitem{baltruvsaitis2016openface}
T.~Baltru{\v{s}}aitis et~al.
\newblock Openface: an open source facial behavior analysis toolkit.
\newblock In {\em WACV}, pages 1--10, 2016.

\bibitem{breiman2001random}
L.~Breiman.
\newblock Random forests.
\newblock {\em Machine learning}, 45(1):5--32, 2001.

\bibitem{canavan2015landmark}
S.~Canavan et~al.
\newblock Landmark localization on 3d/4d range data using a shape index-based
  stat shape model with global and local constraints.
\newblock {\em CVIU}, 139:136--148, 2015.

\bibitem{cootes2001active}
T.~Cootes et~al.
\newblock Active appearance models.
\newblock {\em IEEE Transactions on PAMI}, (6):681--685, 2001.

\bibitem{corneanu2016survey}
C.~Corneanu et~al.
\newblock Survey on rgb, 3d, thermal, and multimodal approaches for facial
  expression recognition: History, trends, and affect-related applications.
\newblock {\em IEEE transactions on PAMI}, 38(8):1548--1568, 2016.

\bibitem{cosker2011facs}
D.~Cosker, E.~Krumhuber, and A.~Hilton.
\newblock A facs valid 3d dynamic action unit database with applications to 3d
  dynamic morphable facial modeling.
\newblock In {\em ICCV}, 2011.

\bibitem{cutler2007random}
D.~Cutler et~al.
\newblock Random forests for classification in ecology.
\newblock {\em Ecology}, 88(11):2783--2792, 2007.

\bibitem{drira20123d}
H.~Drira and others.
\newblock 3d dynamic expression recognition based on a novel deformation vector
  field and random forest.
\newblock In {\em ICPR}, pages 1104--1107, 2012.

\bibitem{ekman1989argument}
P.~Ekman.
\newblock The argument and evidence about universals in facial expressions.
\newblock {\em Handbook of social psychophysiology}, pages 143--164, 1989.

\bibitem{ekman1978Facs}
P.~Ekman and W.~Friesen.
\newblock The facial action coding system: A technique for the measurement of
  facial movement.
\newblock {\em Consulting Psychologists Press}, 1978.

\bibitem{FACS}
P.~Ekman and E.~Rosenberg.
\newblock What the face reveals: Basic and applied studies of spon exp using
  the facial action coding system (facs).
\newblock {\em Oxford Uni Press}, 1997.

\bibitem{fabianoICIP2018}
D.~Fabiano and S.~Canavan.
\newblock Spon and non-spontaneous 3d facial expression recognition using a
  statistical model with global and local constraints.
\newblock {\em ICIP}, 2018.

\bibitem{fanelli20103}
G.~Fanelli et~al.
\newblock A 3-d audio-vis corpus of affect comm.
\newblock {\em IEEE Trans on Multimedia}, 12(6):591--598, 2010.

\bibitem{fanelli2011real}
G.~Fanelli, J.~Gall, and L.~Van~Gool.
\newblock Real time head pose estimation with random regression forests.
\newblock In {\em CVPR 2011}, pages 617--624. IEEE, 2011.

\bibitem{fang20123d}
T.~Fang et~al.
\newblock 3d/4d facial expression analysis: An advanced annotated face model
  approach.
\newblock {\em IVC}, 30(10):738--749, 2012.

\bibitem{friesenEmFacs}
W.~Friesen and P.~Ekman.
\newblock Emfacs-7: Emotional facial action coding system.
\newblock {\em Uni of Cali at SF}, 2(36):1.

\bibitem{hindujafusion}
S.~Hinduja et~al.
\newblock Fusion of hand-crafted and deep features for empathy prediction.
\newblock {\em FG Workshops}, 2019.

\bibitem{kessous2010multimodal}
L.~Kessous et~al.
\newblock Multimodal emotion recognition in speech-based interaction using
  facial expression, body gesture and acoustic analysis.
\newblock {\em Journal on Multimodal User Interfaces}, 3(1-2):33--48, 2010.

\bibitem{knapp2011physiological}
R.~Knapp et~al.
\newblock Phys signals and their use in augmenting emotion recognition for
  human--machine interaction.
\newblock In {\em Emotion-oriented systems}, pages 133--159. 2011.

\bibitem{koelstra2011deap}
S.~Koelstra et~al.
\newblock Deap: A database for emotion analysis; using physiological signals.
\newblock {\em IEEE Transactions on Affective Computing}, 3(1):18--31, 2011.

\bibitem{liu2014facial}
P.~Liu et~al.
\newblock Facial expression recognition via a boosted deep belief network.
\newblock In {\em CVPR}, pages 1805--1812, 2014.

\bibitem{lucey2010extended}
P.~Lucey et~al.
\newblock The extended cohn-kanade dataset (ck+): A complete dataset for action
  unit and emotion-specified expression.
\newblock In {\em CVPRW}, pages 94--101. IEEE, 2010.

\bibitem{mckeown2011semaine}
G.~McKeown et~al.
\newblock The semaine database: Annotated multimodal records of emotionally
  colored conversations between a person and a limited agent.
\newblock {\em IEEE Transactions on Affective Computing}, 3(1):5--17, 2011.

\bibitem{mirza2014conditional}
M.~Mirza and S.~Osindero.
\newblock Conditional generative adversarial nets.
\newblock {\em arXiv preprint arXiv:1411.1784}, 2014.

\bibitem{picard2001toward}
R.~W. Picard, E.~Vyzas, and J.~Healey.
\newblock Toward machine emotional intelligence: Analysis of affective
  physiological state.
\newblock {\em IEEE Trans on PAMI}, (10):1175--1191, 2001.

\bibitem{prkachin2008structure}
P.~Prkachin, K.and~Solomon.
\newblock The structure, reliability and validity of pain expression: Evidence
  from patients with shoulder pain.
\newblock {\em Pain}, 139(2):267--274, 2008.

\bibitem{soleymani2011multimodal}
M.~Soleymani, M.~Pantic, and T.~Pun.
\newblock Multimodal emotion recognition in response to videos.
\newblock {\em IEEE transactions on affective computing}, 3(2):211--223, 2011.

\bibitem{stratou2012exploring}
G.~Stratou et~al.
\newblock Exploring the effect of illumination on automatic expression
  recognition using the ict-3drfe database.
\newblock {\em IVC}, 30(10):728--737, 2012.

\bibitem{SunTrackingVertFlow}
Y.~Sun et~al.
\newblock Tracking vertex vlow and model adaptation for 3d spatio-temporal face
  analysis.
\newblock {\em IEEE Transactions on SMC-A}, 40(3):461--474, 2010.

\bibitem{wang2010natural}
S.~Wang et~al.
\newblock A natural visible and infrared facial expression database for
  expression rec and emotion inference.
\newblock {\em IEEE Trans on Multimedia}, 12(7):682--691, 2010.

\bibitem{yang2018facial}
H.~Yang et~al.
\newblock Facial expression recognition by de-expression residue learning.
\newblock In {\em CVPR}, 2018.

\bibitem{yang2018identity}
H.~Yang et~al.
\newblock Identity-adaptive facial expression recognition through expression
  regeneration using conditional generative adversarial networks.
\newblock In {\em FG}, 2018.

\bibitem{yin126high}
L.~Yin et~al.
\newblock A high-resolution 3d dynamic facial expression database.
\newblock In {\em FG}, volume 126, 2008.

\bibitem{zamzmi2016machine}
G.~Zamzmi et~al.
\newblock Machine-based multimodal pain assessment tool for infants: a review.
\newblock {\em arXiv preprint arXiv:1607.00331}, 2016.

\bibitem{zhang2016multimodal}
Z.~Zhang et~al.
\newblock Multi spon emotion corpus for human behavior analysis.
\newblock In {\em CVPR}, pages 3438--3446, 2016.

\bibitem{zhao2012real}
X.~Zhao et~al.
\newblock Real-time hand gesture detection and recognition by random forest.
\newblock In {\em Communications and information processing}, pages 747--755.
  2012.

\end{thebibliography}
